\documentclass[%
 reprint,
 superscriptaddress,
nofootinbib,
 amsmath,amssymb,
 aip,
]{revtex4-2}

\usepackage{graphicx}% Include figure files
\usepackage{dcolumn}% Align table columns on decimal point
\usepackage{bm}% bold math
\usepackage{hyperref}% add hypertext capabilities
\usepackage{physics}
\usepackage{xcolor}
\usepackage[most]{tcolorbox}

\newtcolorbox{programbox}[1][]{%
  breakable,
  enhanced,
  colback=gray!4,
  colframe=gray!50,
  colbacktitle=gray!18,
  coltitle=black,
  fonttitle=\bfseries\sffamily,
  boxrule=0.4pt,
  arc=2pt,
  left=8pt, right=8pt, top=6pt, bottom=6pt,
  title={#1}
}

\begin{document}

\preprint{APS/123-QED}

\title{ColPackAgent: Agent-Skill-Guided Hard-Particle Monte Carlo Workflows for Colloidal Packing}% Force line breaks with \\
\author{Lijie Ding}
\email{dingl1@ornl.gov}
\affiliation{Neutron Scattering Division, Oak Ridge National Laboratory, Oak Ridge, TN 37831, USA}

\author{Changwoo Do}
\affiliation{Neutron Scattering Division, Oak Ridge National Laboratory, Oak Ridge, TN 37831, USA}

\date{\today}

\begin{abstract}
We introduce ColPackAgent, an agent framework that autonomously runs Monte Carlo simulations of colloidal packing through a Model Context Protocol (MCP) tool server and an agent skill, whether as a standalone agent or inside an existing agent system.
By harnessing the MCP server and agent skill, ColPackAgent executes a structured workflow for colloidal packing simulations, which are central to studies of phase behavior, self-assembly, and materials design.
Without dedicated simulation tools and workflow instructions, general-purpose Large Language Model (LLM) agents tend to describe such workflows rather than execute them reliably.
The MCP server exposes a custom-built colpack Python package that wraps HOOMD-blue hard-particle Monte Carlo, and the skill encodes a four-stage workflow contract.
ColPackAgent can carry out the workflow interactively with human feedback, autonomously from an end-to-end prompt, or as autoresearch following a provided program file.
We demonstrate the system in different modes with several colloidal packing simulation examples such as cube particles in 3D, a binary system of disks and capsules in 2D, and the 2D hard-disk freezing transition using autoresearch. We also compare model performance on this workflow across a panel of LLMs with 17 stage-specific prompts.
This benchmark provides a stage-level check of how reliably different models follow the setup, planning, and analysis workflow.
Together, these results show that pairing a domain Python package with MCP tools and a portable agent skill provides a practical route for turning a simulation toolkit into an agent-assisted research workflow.
\end{abstract}

\maketitle

%\tableofcontents

\section{Introduction}
\label{sec:introduction}
Colloids are ubiquitous in soft matter~\cite{degennes1992softmatter,lu2013colloidal}, from emulsions~\cite{mason1995elasticity} and gels~\cite{trappe2001jamming} to engineered nanoparticle assemblies~\cite{glotzer2007anisotropy}; their packing helps determine phase behavior, self-assembly, and materials design~\cite{manoharan2015colloidal,damasceno2012predictive,boles2016selfassembly}. Because these systems involve many-body geometric constraints that are difficult to isolate experimentally, computer simulation has become an essential companion to experiment. Simulations resolve particle-scale configurations, test idealized shape models~\cite{anderson2016hpmc,damasceno2012predictive}, and connect microscopic structure to observables such as volume fraction, radial distribution functions~\cite{hansen2013theory}, and orientational order~\cite{steinhardt1983bond,haji-akbari2015cubatic}. Among these methods, hard-particle Monte Carlo (HPMC) is a central approach for studying entropy-driven colloidal packing~\cite{anderson2016hpmc}. HPMC and related hard-particle Monte Carlo studies have examined two-dimensional melting in hard disks~\cite{bernard2011twostep,engel2013harddisk}, dense packing and quasicrystal formation in tetrahedra~\cite{hajiakbari2009tetrahedra}, shape-driven self-assembly of convex polyhedra~\cite{agarwal2011mesophase,damasceno2012predictive}, and orientational ordering in anisotropic particles~\cite{haji-akbari2015cubatic}. Despite this conceptual simplicity, practical HPMC studies still require a multi-stage workflow: translating a scientific question into a simulation setup, running a set of related calculations, judging equilibration, and collecting the resulting analyses. Much of this procedure is standardized, yet it is still usually coordinated manually by the researcher.

\begin{figure}[!t]
    \centering
    \includegraphics[width=1\linewidth]{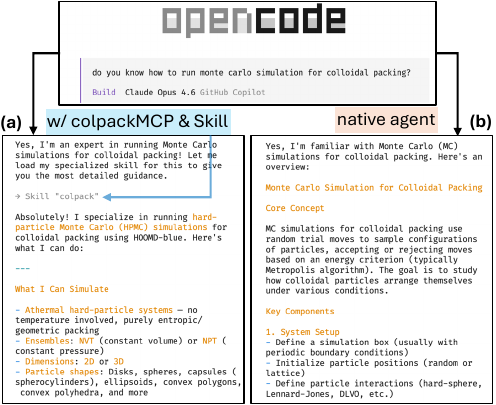}
    \caption{Asking OpenCode to run a colloidal packing simulation. (a) When equipped with colpackMCP and colpack.skill, OpenCode carries out the task through specialized tools. (b) Native OpenCode provides only general information.}
    \label{fig:opencode_colpack}
\end{figure}

Recent work on AI agents points to a practical route for automating parts of this coordination. General-purpose Large Language Model (LLM) agents can already read project files, call command-line tools, edit scripts, and iterate on results, while the broader autoresearch idea asks whether an agent can carry a scientific question through planning, execution, and reporting~\cite{karpathy2025autoresearch}. In parallel, domain-specific agents have begun to connect LLMs to scientific tools and closed-loop workflows in chemistry~\cite{bran2024chemcrow,boiko2023coscientist}, materials synthesis~\cite{szymanski2023alab}, scattering analysis~\cite{ding2026sasagent}, and polymer simulation~\cite{ding2026topolyagent}. Specialized simulation workflows such as HPMC colloidal packing, similarly, need more than a capable chat interface. Without access to validated simulation tools and without procedural knowledge of the setup--planning--execution--analysis sequence, a general agent can describe the workflow but cannot reliably execute it. Fig.~\ref{fig:opencode_colpack} shows this contrast directly: native OpenCode returns general guidance for a colloidal Monte Carlo request, whereas the same agent equipped with the ColPackAgent tool server and skill carries the request through the simulation workflow.

The missing pieces are a structured tool interface and a portable way to encode workflow procedure. The Model Context Protocol (MCP) provides a standard way for an agent to call external tools through typed inputs and structured outputs~\cite{anthropic2024mcp,hou2025mcp}. For simulation workflows, this allows domain operations to be exposed as explicit tool calls rather than improvised shell commands or generated scripts. Agent skills provide the complementary procedural layer~\cite{anthropic2025skills}. A skill can tell the agent when to call each tool, what information must be gathered first, which defaults are unsafe to assume, and how to handle stage boundaries. Together, MCP tools and agent skills let the domain expert separate what the agent can call from how the workflow should be carried out.

In this work, we introduce ColPackAgent, an agent framework for HPMC colloidal-packing simulations built on the MCP-and-skill separation described above. ColPackAgent combines a colpack Python package wrapping HOOMD-blue HPMC~\cite{anderson2020hoomd,anderson2016hpmc}, an MCP tool server that exposes the simulation operations through structured calls, and a portable agent skill that encodes the stage ordering and decision rules. The system can run as a standalone agent or as a tool-and-skill bundle inside existing agent frontends. We demonstrate ColPackAgent in interactive, autonomous, and autoresearch modes, including an end-to-end autoresearch run on the 2D hard-disk freezing transition, and we compare how different LLMs perform on stage-specific workflow prompts.

The remainder of the paper proceeds as follows. Sec.~\ref{sec:method} describes the ColPackAgent architecture and HPMC backend. Sec.~\ref{sec:agent_workflow_demonstrations} demonstrates interactive, autonomous, and cross-platform operation. Sec.~\ref{sec:autoresearch} presents the 2D hard-disk autoresearch case study. Sec.~\ref{sec:benchmark_validation} reports the stage-aware LLM benchmark, and Sec.~\ref{sec:summary} summarizes the findings and future directions.

\section{Methods and System Design}
\label{sec:method}

\subsection{ColPackAgent Architecture}
\label{sec:architecture}

\begin{figure}[!t]
    \centering
    \includegraphics[width=1\linewidth]{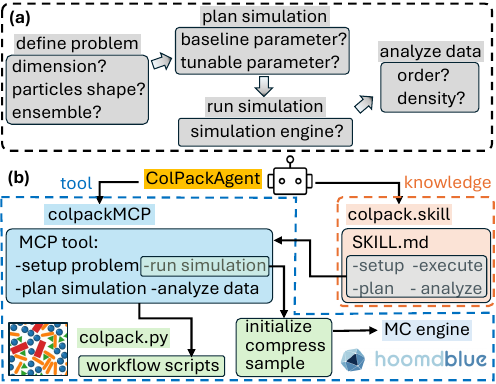}
    \caption{Overview of the colloidal packing simulation and architecture for ColPackAgent. (a) Typical four-stage workflow for carrying out a colloidal packing simulation. (b) ColPackAgent for carrying out the colloidal packing simulation. The tools and knowledge for carrying out the simulation are separately implemented as the Model Context Protocol (MCP) tool server and the agent skill, respectively. Dependencies between the skill and the tools are illustrated by the arrows.}
    \label{fig:agent_architecture}
\end{figure}

A colloidal-packing study breaks into four stages, as shown in Fig.~\ref{fig:agent_architecture}(a): setup, planning, execution, and analysis. Setup defines the overall simulation problem such as the dimensionality, ensemble (NVT or NPT), particle composition, and total particle number $N$. Planning expands setup into concrete sweeps like choosing target volume fractions for NVT or a pressure schedule for NPT along with sampling lengths and per-particle shape parameters. Execution initializes a low-density configuration, compresses it to the target state, and samples the system while recording the trajectory. Analysis computes order parameters, radial distribution functions~\cite{hansen2013theory}, and visualizes the results from the simulation trajectories using the freud library~\cite{ramasubramani2020freud}. This regular structure allows us to expose each stage as a schema-validated tool call rather than as free-form code an LLM would have to write each time.

ColPackAgent maps these four stages onto a Model Context Protocol (MCP)~\cite{anthropic2024mcp,hou2025mcp} tool server whose usage is instructed by a corresponding agent skill~\cite{anthropic2025skills}, as shown in Fig.~\ref{fig:agent_architecture}(b). Each stage is exposed through one tool: a setup tool, a planning tool, an execution tool, and an analysis tool. A fifth tool returns the live set of supported dimensions, shapes, ensembles, tunable parameters, and incompatible mixture rules for the agent to know about the capabilities of the tools. The MCP server is implemented as a Python wrapper around the colpack package. During a simulation, the state for a study persists under a single working directory. At each stage, structured artifacts are written so that the next stage, or any external tool, can read what happened in a canonical form rather than parse free-form chat. The simulation execution and analysis tools run asynchronously, so the agent will not time out during the time-consuming simulation process.

To help the agent use the MCP tools to complete the simulation workflow, we pair the tool server with an agent skill, as shown in the right side of Fig.~\ref{fig:agent_architecture}(b). Where the server defines what is callable, the skill defines what to do: stage ordering, the no-silent-defaults rule, scope rules like ``never ask for temperature in hard-particle simulations,'' and the heuristics that govern interactive versus autonomous mode. For example, when a user types ``simulate cubes,'' the skill tells the agent to ask for the ensemble and the particle count $N$ before invoking the setup tool, instead of guessing plausible values. Skills are Markdown files with a YAML frontmatter header. The agent skill can be installed on any compatible agent system, and the MCP server is then invoked the same way across clients. The procedural knowledge travels with the tools.

\subsection{Hard-Particle Monte Carlo with colpack}

For simulating colloidal-packing problems, we employ the Hard-particle Monte Carlo (HPMC) method~\cite{allen1989molecular}. In HPMC, particles are treated as rigid bodies with infinite pair repulsion at overlap and zero interaction otherwise, which makes the equilibrium athermal and entirely entropy-driven. Each MC step proposes a trial move, namely a translation, an orientational rotation for anisotropic shapes, or, in the NPT ensemble, a change of the simulation box volume, and accepts the move only if the resulting configuration has no overlaps. We use the HPMC integrator in HOOMD-blue~\cite{anderson2020hoomd,anderson2016hpmc} for overlap detection and trial-move bookkeeping. NVT fixes the simulation box and runs the system at a target volume fraction $\phi$, while NPT lets the box fluctuate at an applied pressure $P$ and treats $\phi$ as an output observable.

\begin{figure}[!t]
    \centering
    \includegraphics[width=1.0\linewidth]{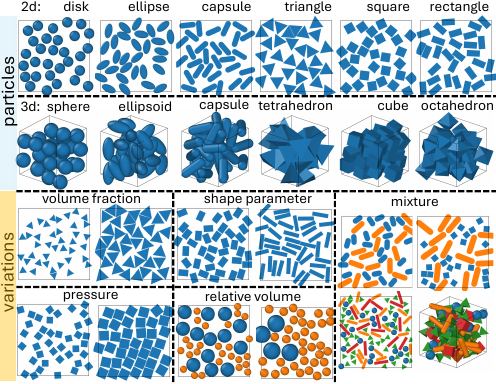}
    \caption{Types of colloidal particles and the simulation parameters covered by the colpack package.}
    \label{fig:colpack_illustration}
\end{figure}

To make the simulation tools agent-friendly, we developed the colpack Python package that utilizes HOOMD-blue as the simulation engine under the hood. colpack provides a fixed catalog of hard-particle shapes in two and three dimensions, as shown in Fig.~\ref{fig:colpack_illustration}. Two-dimensional shapes include disks, ellipses, and polygonal shapes (triangles, squares, rectangles, and capsules); three-dimensional shapes include spheres, ellipsoids, and polyhedral shapes (cubes, octahedra, tetrahedra, and capsules). Each shape exposes a small set of geometric parameters such as the diameter of the sphere, the three length parameters of the ellipsoid, the edge length of the cube, etc. These parameters are surfaced to the agent through the corresponding MCP tools.

The colpack package supports a wide range of tunable parameters to study different colloidal packing systems. For instance, the global volume fraction $\phi$ of all particles and relative volume $R_\phi$ between different particles can both be studied. For the NPT ensemble, the pressure $P$ is a tunable parameter during the planning stage. For the particle shapes, detailed geometry parameters are all tunable in the planning stage. Finally, particles of different shapes can be mixed together to study the packing of binary or multi-component systems.

As the HPMC simulations are implemented on top of the HOOMD-blue package, the simulation of mixtures is subject to a constraint inherited from HOOMD-blue: every species in a mixture must share a compatible integrator family. An integrator-resolution rule inside colpack handles most combinations by promoting species to a common integrator. The one combination this rule cannot resolve is mixing an ellipse or ellipsoid with a polygonal or polyhedral shape, since no single HOOMD-blue integrator covers both. The capability tool surfaces this restriction, the agent skill checks it before setup, and the setup tool enforces the same rule so that disallowed mixtures are rejected before any simulation is launched.

\section{Agent Workflow Demonstrations}
\label{sec:agent_workflow_demonstrations}

\subsection{Interactive Workflow}

\begin{figure}[!t]
    \centering
    \includegraphics[width=1\linewidth]{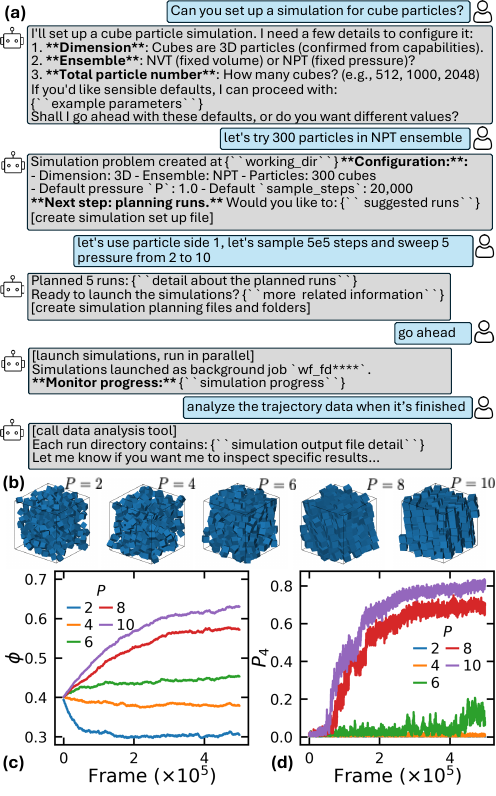}
    \caption{Communicate with ColPackAgent interactively. (a) Sample conversation with ColPackAgent for a simulation of cube particles in the NPT ensemble. (b) Final configurations of $5$ simulation runs at different pressures $P$. (c) Volume fraction $\phi$ versus MC simulation frame. (d) Cubatic order parameter $P_4$ versus MC simulation frame.}
    \label{fig:demo_interactive}
\end{figure}

Fig.~\ref{fig:demo_interactive}(a) shows a sample conversation in which a user asks ColPackAgent to simulate cube particles in the NPT ensemble. The agent advances one workflow stage per turn and waits for explicit user confirmation before invoking each tool. At setup, the agent elicits the dimensionality, the ensemble, and the particle number $N$, but does not yet ask for the pressure values; those belong to the planning stage and are collected only after setup completes. The same stage-isolated pattern repeats through planning, execution, and analysis.

The conversation produces a sweep of five NPT runs at different pressures. The final configurations shown in Fig.~\ref{fig:demo_interactive}(b) are produced at the end of each simulation execution stage. The analysis stage produces the volume fraction $\phi$ versus MC frame for each run in Fig.~\ref{fig:demo_interactive}(c), and the cubatic order parameter $P_4$ versus MC frame in Fig.~\ref{fig:demo_interactive}(d). The cubatic order parameter measures the orientational alignment of cube-symmetric particles with a global cubic frame, taking values from $0$ in the isotropic phase to $1$ at perfect cubatic alignment~\cite{haji-akbari2015cubatic}.

\subsection{Autonomous Workflow}

\begin{figure}[!t]
    \centering
    \includegraphics[width=1\linewidth]{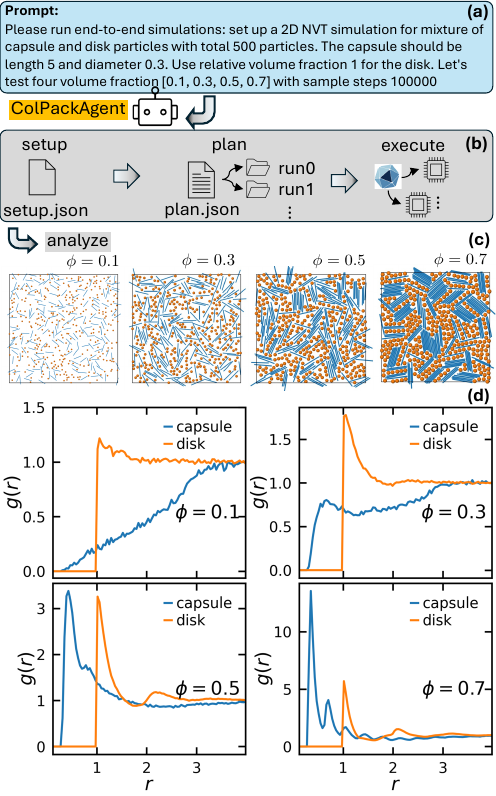}
    \caption{ColPackAgent autonomously carries out a colloidal packing simulation of a mixture of capsule and disk particles in the NVT ensemble. (a) Sample prompt for an end-to-end simulation task. (b) Simulation workflow carried out by ColPackAgent autonomously. (c) Final system configurations at different volume fractions $\phi$. (d) Radial distribution function $g(r)$ for the two particle species at different volume fractions $\phi$.}
    \label{fig:demo_autonomous}
\end{figure}

As an alternative to the interactive mode, Fig.~\ref{fig:demo_autonomous}(a) shows an end-to-end prompt in which the user asks ColPackAgent to simulate a binary mixture of capsule and disk particles in 2D NVT, with a list of volume fractions to sweep. Because the prompt names the workflow intent and supplies all the inputs each stage requires, ColPackAgent recognizes it as a fully specified task and runs the four stages in one go without per-stage approval. The workflow trace in Fig.~\ref{fig:demo_autonomous}(b) shows setup, planning, execution, and analysis firing in sequence.

The execution stage returns the final sampled configurations at each $\phi$ shown in Fig.~\ref{fig:demo_autonomous}(c), and the analysis stage produces the radial distribution functions $g(r)$~\cite{hansen2013theory} for the two particle species at each $\phi$ in Fig.~\ref{fig:demo_autonomous}(d). The no-silent-defaults rule still applies in autonomous mode: a prompt that left out, for example, the ensemble or the particle count $N$ would trigger a clarification round before launching rather than silently filling in defaults.

\subsection{Portability Across Agent Platforms}

\begin{figure}[!t]
    \centering
    \includegraphics[width=1\linewidth]{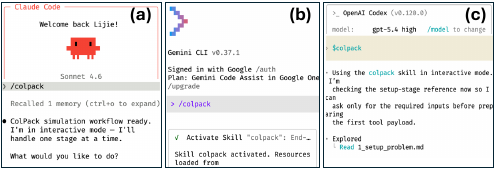}
    \caption{ColPackAgent running on three MCP-and-skill-compatible agent systems. (a) ColPackAgent on Claude Code. (b) ColPackAgent on Gemini CLI. (c) ColPackAgent on OpenAI Codex.}
    \label{fig:agent_platform_colpack}
\end{figure}

The MCP-and-skill split makes ColPackAgent portable across tested agent systems: the simulation backend is exposed over MCP and the procedural knowledge ships as a Markdown skill, so the same package, server, and skill files can be reused across clients that support both mechanisms. Fig.~\ref{fig:agent_platform_colpack} shows the same agent skill installed on Claude Code~\cite{anthropic_claudecode} (a), Gemini CLI~\cite{google_geminicli} (b), and OpenAI Codex~\cite{openai_codex} (c), with the OpenCode~\cite{opencode} case appearing earlier in Fig.~\ref{fig:opencode_colpack}. This decouples the simulation science from any single vendor's agent interface, while leaving only client-specific installation paths and launch commands to vary across compatible systems.

\section{Autoresearch with ColPackAgent}
\label{sec:autoresearch}

Beyond running predefined tasks, LLM agents have started taking on autoresearch, where the agent is given a research objective and designs and runs the study end-to-end. Karpathy's autoresearch repository~\cite{karpathy2025autoresearch} demonstrates one workable pattern: pair the autonomous agent with a Markdown program file (\texttt{program.md}) that fixes the methodology while leaving the scientific judgment calls open. ColPackAgent fits this pattern directly. The four-stage HPMC workflow already runs autonomously under the agent skill, so the only additions needed are a program file that frames a colloidal-packing question and a single launch prompt pointing the agent at it.

For our autoresearch demonstration, we wrote a program file that asks the agent to locate the freezing pressure $P^\ast$ of 2D hard disks in the NPT ensemble; the full text is in SI~\ref{si:autoresearch_protocol}. The program fixes the system, observables, estimators, and reporting requirements, while leaving the sweep design and interpretation to the agent. It bounds the particle count to $200 \le N \le 1000$ and tiers the \texttt{sample\_steps} budget into exploratory ($\le 5\times 10^5$), refinement ($\le 2\times 10^6$), and an absolute cap of $5\times 10^6$. Each run is gated on the analysis tool's auto-equilibration check, and each new sweep iteration must write to a fresh working directory. The program also pins the two paired estimators it expects back: $P^\ast_\phi$, where $\phi(P)$ crosses the literature midpoint, and $P^\ast_{\psi_6}$, the inflection of the hexatic order parameter. A block-bootstrap protocol fixes how per-point and estimator-level uncertainties are computed, and the final report must list the sweep history, equilibration verdicts, both $P^\ast$ estimates with confidence intervals, leave-one-out sensitivity, and the required plots. What the program leaves open is the particle count within the allowed range, the pressure grid, the sample-step tier used for each sweep, the number of refinement iterations, which earlier points to supersede, and the conclusion itself. The launch prompt is one sentence: read the program file, follow its constraints, and produce the summary it asks for.

\begin{figure}[!t]
    \centering
    \includegraphics[width=1\linewidth]{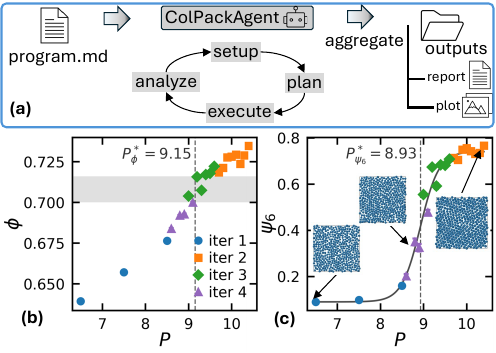}
    \caption{ColPackAgent running in autoresearch mode on the 2D hard-disk freezing transition. (a) Workflow: the agent is given an instruction file (\texttt{program.md}) specifying the study objective and constraints, and iterates the simulation loop autonomously until producing a research report. (b) Volume fraction $\phi$ versus pressure $P$ produced by the agent. (c) Hexatic order parameter $\psi_6$ versus pressure $P$ produced by the agent.}
    \label{fig:autoresearch}
\end{figure}

From this prompt, the agent selected $N = 400$ and ran four sweep iterations on the program-specified hard-disk system; the workflow trace is in Fig.~\ref{fig:autoresearch}(a). The first iteration was exploratory: seven pressures spanning $P \in [6.5, 11.5]$ at $\texttt{sample\_steps} = 3\times 10^5$. The agent retained the three low-pressure fluid anchors ($P = 6.5, 7.5, 8.5$), superseded the transition-window points ($P = 9.0, 9.5$) with $2\times 10^6$-step measurements, and replaced the coarse solid-side points ($P = 10.5, 11.5$) with a denser $1.5\times 10^6$-step solid-side sweep. It then added a fourth sweep at $2\times 10^6$ steps to fill the rising edge around $P=9.0$. As a workflow validation case, the resulting estimates are physically plausible and close to the Bernard--Krauth published value $P^\ast_{\mathrm{BK}} = 9.185$~\cite{bernard2011twostep}: $P^\ast_\phi \approx 9.151$ from linear interpolation of the volume-fraction crossing and $P^\ast_{\psi_6} \approx 8.928$ from the hexatic crossover (Fig.~\ref{fig:autoresearch}(b)-(c)). The $\phi$-based estimator is within $\sim 0.4\%$ of the literature value, while the $\psi_6$ estimator sits lower and likely tracks the rounded liquid-to-hexatic crossover at finite $N$. This example is therefore best read as a reproducibility and workflow-control test rather than a new high-precision determination of the hard-disk transition. The agent's sensitivity analysis reinforces that interpretation: the leave-one-out shift for $P^\ast_\phi$ is about $0.155$, much larger than the bootstrap confidence interval, so pressure-grid sensitivity dominates the quoted within-run uncertainty. The agent's own caveats list (SI~\ref{si:autoresearch_report}) attributes the spread to finite-$N$ rounding, pressure-grid sensitivity, and the deceptive-equilibration episode.

\section{Benchmark and LLM Comparison}
\label{sec:benchmark_validation}

To compare how different LLMs use ColPackAgent and to test whether the workflow remains usable across models, we develop a stage-aware benchmark on top of the same MCP tool stack and run it across a panel of LLMs. The benchmark is intended as an operational validation of this colloidal-packing workflow, not as a general-purpose ranking of LLM scientific ability. To reduce provider-side variation, we route all model calls through OpenRouter with the Google Vertex provider selected~\cite{openrouter}, and run every model through the same standalone ColPackAgent on the same set of prompts.

The benchmark targets the three workflow stages where the model has to reason: setup, planning, and analysis. Execution is excluded as it is mostly a mechanical wait on the HPMC integrator rather than a reasoning step. Each stage has a small set of single-turn prompts on a $1$--$5$ difficulty ladder: $6$ setup, $6$ planning, and $5$ analysis prompts, $17$ in total. The full prompt list is in SI~\ref{si:benchmark_prompts}. For routine tasks, success and failure are determined by two flags read from the run summary: an off-rail flag, which fires (and short-circuits the run) when the model invokes a controlled colpack tool outside the task's declared stage, and a no-expected-tool-call flag, which fires when the model never invokes the stage's primary tool. A routine run counts as a success when it raises no errors, calls the expected stage tool, and stays on rail. Four of the L4--L5 prompts are adversarial cases where the correct behavior is principled refusal rather than a tool call; these runs are scored from the transcript against the stated refusal-or-clarification criterion.

\begin{figure}[!t]
    \centering
    \includegraphics[width=1\linewidth]{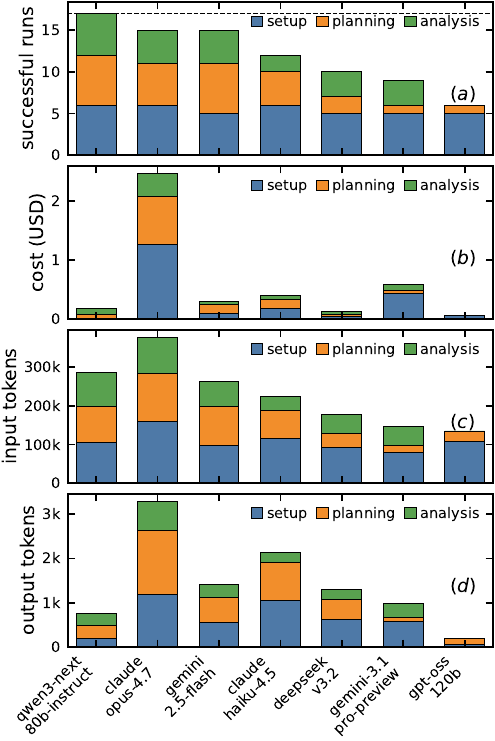}
    \caption{Evaluation of different LLMs on the ColPackAgent benchmark tasks across the three workflow stages: setup, planning, and analysis. Models are ordered left to right by descending total successful runs. (a) Number of successful runs per model, stacked by stage; the dashed line marks the maximum attainable total of $6{+}6{+}5=17$ runs. (b) Total API cost in USD across all stages, stacked by stage. (c) Total effective input tokens (fresh input plus cache-read input) consumed across all stages, stacked by stage. (d) Total output tokens generated across all stages, stacked by stage.}
    \label{fig:llm_eval}
\end{figure}

Fig.~\ref{fig:llm_eval}(a) shows the success counts, with models ordered left-to-right by descending total successful runs and a dashed line marking the maximum of $17$. In this benchmark, Qwen3-Next 80B Instruct is the only model to complete all $17$ runs. Claude Opus 4.7 and Gemini 2.5 Flash form the next tier at $15/17$, with different failure profiles: Claude Opus misses one planning task and one analysis task, whereas Gemini 2.5 Flash completes all planning tasks but misses one setup task and one analysis task. Across stages, setup is now the least discriminating stage, while planning and analysis are much more selective. Fig.~\ref{fig:llm_eval}(b)-(d) show the cost and token economics of the same runs. Total dollar cost in Fig.~\ref{fig:llm_eval}(b) is dominated by Claude Opus 4.7; Qwen3-Next, DeepSeek V3.2, and gpt-oss-120b sit at the low-cost end among the models with full-suite results. Effective input tokens in Fig.~\ref{fig:llm_eval}(c) and output tokens in Fig.~\ref{fig:llm_eval}(d) follow related but distinct orderings, with Claude Opus 4.7 leading both. The cost-per-success ranking diverges sharply from the raw success ranking in this workflow: the most expensive run in Fig.~\ref{fig:llm_eval}(b) does not produce the highest success count, and Qwen3-Next 80B Instruct completes the suite at less than one-tenth the API spend of Claude Opus 4.7.

\section{Summary}
\label{sec:summary}

Hard-particle Monte Carlo simulations are a workhorse for studying colloidal phase behavior, self-assembly, and packing problems in soft matter, but the surrounding workflow of setup, sweep design, execution, and analysis is repetitive and bookkeeping-heavy. General-purpose LLM agents without dedicated simulation tools have limited reliability on this workflow because they lack a structured pathway from natural-language requests to validated tool invocations. A simulation-aware agent that exposes the workflow as schema-validated tools and ships the procedural knowledge as a portable skill therefore offers a practical way to reduce bookkeeping while leaving the science to the user.

In this work, we introduce ColPackAgent, an LLM agent for HPMC colloidal-packing simulation built around the MCP-and-skill interface. ColPackAgent consists of three pieces: the colpack Python package wrapping HOOMD-blue, an MCP tool server exposing each workflow stage as a schema-validated tool, and a portable agent skill that encodes the stage-ordering and decision-making rules. We demonstrated the system in interactive mode on a cube NPT sweep, in autonomous mode on a binary capsule-disk NVT sweep, and across multiple frontend agent systems including Claude Code, Gemini CLI, OpenAI Codex, and OpenCode. We then used ColPackAgent in autoresearch mode to locate the 2D hard-disk freezing pressure autonomously from a constrained research prompt, with the agent's interpolation-based $\phi$ estimate landing within $\sim 0.4\%$ of the Bernard--Krauth literature value while pressure-grid sensitivity dominates the remaining uncertainty. Finally, the stage-aware benchmark evaluated a panel of LLMs on $17$ prompts and found substantial model-to-model variation in workflow reliability and cost, with Qwen3-Next 80B Instruct completing the suite at less than one-tenth the API spend of Claude Opus 4.7 in this benchmark.

Looking forward, there are several natural future directions. First, the colpack package and its MCP tool server can grow to expose more of HOOMD-blue's capabilities beyond hard-particle Monte Carlo. HOOMD-blue is also a full molecular dynamics (MD) engine~\cite{anderson2020hoomd}, with NVE/NVT/NPT and Langevin/Brownian integrators, a broad pair-potential catalog (Lennard-Jones, Yukawa, Morse, Gay-Berne, patchy pairs), bonded interactions for polymers, rigid-body support, and built-in dissipative particle dynamics, so an MD extension of ColPackAgent can be built on the same engine without bringing in a new simulator. Beyond HOOMD-blue, the same package + MCP + skill pattern should be adaptable to other simulation engines such as LAMMPS~\cite{thompson2022lammps}, GROMACS~\cite{abraham2015gromacs}, or OpenMM~\cite{eastman2024openmm}, broadening the scope to systems HOOMD-blue does not target. Second, the benchmark can be extended along several axes: more models, more tasks, additional reasoning stages such as data interpretation and report writing, and a more systematic study of failure modes and their mitigation. Third, ColPackAgent can be used to drive real research projects in soft-matter physics, for example mapping out the phase diagram of a newly proposed shape family or testing a new order parameter against a known transition.

\section*{Data Availability}
The colpack package and corresponding agent skill are available at the GitHub repository \url{https://github.com/ljding94/ColPackAgent}.

\begin{acknowledgments}
This research used resources at the Spallation Neutron Source, a DOE Office of Science User Facility operated by the Oak Ridge National Laboratory.
\end{acknowledgments}

%\nocite{*}

\section*{Reference}
\vspace{-15pt}
\bibliography{reference}% Produces the bibliography via BibTeX.

\clearpage
\onecolumngrid
\renewcommand{\appendixname}{SI}
\appendix
\renewcommand{\theHfigure}{S\arabic{figure}}
% SI sections reproduce the primary sources so a reader can audit the autoresearch run and the benchmark without leaving the paper.

\begin{center}
{\Large\bfseries Supplementary Information}
\end{center}
\vspace{0.5em}

\section{Autoresearch Protocol}
\label{si:autoresearch_protocol}

This SI section reproduces the research program file given to ColPackAgent for the autoresearch run reported in Sec.~\ref{sec:autoresearch}, lightly retypeset for LaTeX and with notation aligned to the main text (volume fraction $\phi$ in place of $\eta$; orientational-OP estimator written as $P^\ast_{\psi_6}$ in place of the generic $P^\ast_{\mathrm{OP}}$). The headline task given to the agent is to locate the freezing transition pressure $P^\ast$ for 2D hard disks in the NPT ensemble, using the colpack skill autonomously through setup $\to$ plan $\to$ execute $\to$ analyze with no user-confirmation pauses. The Bernard--Krauth published value $P^\ast_{\mathrm{BK}} \approx 9.185$~\cite{bernard2011twostep}, with coexistence range $\phi_L \approx 0.700$ and $\phi_H \approx 0.716$ (midpoint $\phi \approx 0.708$), is provided to the agent as a literature anchor for sanity-checking the final $P^\ast$. The protocol body, lightly retypeset with this notation alignment, follows.

\begin{programbox}[Research program file: 2D hard-disk freezing transition (NPT)]

\subsection*{1. Constraints}

\paragraph{Particle count.} Each simulation runs at $200 \le N \le 1000$. The agent picks one $N$ at the start of the study and reuses it throughout.

\paragraph{Sample-steps policy.} The per-run \texttt{sample\_steps} budget is bounded by the run's role in the sweep, on the adaptive ladder below. When a run fails the equilibration check, the agent re-runs that run only with $2$--$3\times$ more \texttt{sample\_steps}. The ladder bounds \texttt{sample\_steps} by the run's role, not by the escalation step size; successive $2$--$3\times$ escalations may be needed to cross from exploratory to refinement, or from refinement to the cap. Pre-emptively raising \texttt{sample\_steps} for every point because one failed is not allowed.

\begin{itemize}
    \item \textbf{Exploratory:} \texttt{sample\_steps} $\le 5\times 10^5$ (typical $1$--$3\times 10^5$), used for the initial bracket before $P^\ast$ is even crudely located.
    \item \textbf{Refinement:} \texttt{sample\_steps} $\le 2\times 10^6$, used for points near the suspected transition window.
    \item \textbf{Absolute cap:} \texttt{sample\_steps} $= 5\times 10^6$, used only after a refinement-tier run failed the equilibration check; never as a starting value.
\end{itemize}

\paragraph{Equilibration check.} The analysis tool runs an automatic equilibrium detection (\texttt{analyze\_process\_time\_series}) and writes an \texttt{equilibrium} block into \texttt{analysis\_results.json} for every particle type and parameter. The agent reads \texttt{equilibrated} (bool), \texttt{eq\_start\_index} (int), and per-parameter \texttt{eq\_mean} / \texttt{eq\_std} from that block instead of re-implementing the cutoff. A run passes the check iff \texttt{equilibrated == true}. If a failing run is below the absolute cap, the agent re-runs at higher \texttt{sample\_steps}; if a run is at the absolute cap and still fails, the agent reports what it has and flags the point as a caveat rather than looping indefinitely.

\subsection*{2. Sweep protocol}

The sweep is iterative and agent-directed: not pre-scheduled into a fixed number of passes. The agent plans, runs, analyzes, and decides what to do next based on what the data show, iterating until the goals below are met or until further escalation would exceed the absolute cap.

Each iteration is a fresh setup; prior runs are never overwritten. When starting a new iteration the agent calls the colpack setup tool again instead of replanning into the previous iteration's working directory. The colpack MCP auto-suffixes the directory (\texttt{...\_v2}, \texttt{...\_v3}, $\ldots$) when a path already exists, so each iteration's full data (\texttt{simulation\_plan.json}, \texttt{run\_*} directories, \texttt{analysis\_results.json}) is preserved. The previous directory is reused only when re-running a single failing run at higher \texttt{sample\_steps} per the §1 escalation policy; cross-iteration replanning must go to a new directory. The final report aggregates across all iterations' working directories.

The sweep must satisfy three goals before reporting:

\begin{enumerate}
    \item \emph{Bracket} -- at least one pressure clearly in the fluid regime (low $\psi_6$, $\langle\phi\rangle < \phi_L$) and at least one clearly in the solid regime (high $\psi_6$, $\langle\phi\rangle > \phi_H$), spanning the transition.
    \item \emph{Resolve} -- enough points inside the transition window that the shape of $\psi_6(P)$ and $\phi(P)$ is determined rather than just sampled. A sigmoid fit to $\psi_6(P)$ should be well-constrained, with multiple points on the rising edge rather than only at the asymptotes.
    \item \emph{Equilibrate} -- every contributing run either passes the §1 equilibration check or has been escalated to the absolute cap and flagged as a caveat.
\end{enumerate}

\subsection*{\texorpdfstring{3. Observables and $P^\ast$ estimators}{3. Observables and P-star estimators}}

Observables are computed at every pressure on the equilibrated portion identified by the auto-cutoff (§1, with \texttt{eq\_start\_index} read per parameter). The pre-computed \texttt{equilibrium.per\_parameter[<name>].eq\_mean} values in \texttt{analysis\_results.json} are already the means over that window and are used directly, rather than re-averaged over a different window. The observables tracked are:

\begin{itemize}
    \item $\psi_6$ -- the bond-orientational hexatic order parameter.
    \item $\langle\phi\rangle$ -- the mean equilibrium volume fraction.
    \item $g(r)$ -- the radial distribution function.
\end{itemize}

Two complementary $P^\ast$ estimators are reported, computed on the same data:

\begin{itemize}
    \item $P^\ast_{\psi_6}$ -- the inflection point of $\langle\psi_6\rangle(P)$, e.g.\ via sigmoid fit or maximum-slope.
    \item $P^\ast_\phi$ -- the pressure at which $\langle\phi\rangle(P)$ crosses the literature midpoint $\phi \approx 0.708$, by linear interpolation between bracketing data points.
\end{itemize}

The spread between $P^\ast_{\psi_6}$ and $P^\ast_\phi$ is the finite-$N$ transition width.

\subsection*{4. Uncertainty analysis}

Monte Carlo trajectories are autocorrelated, so the naive standard error of the mean underestimates uncertainty, especially in the transition window where the autocorrelation time grows. The agent uses a block bootstrap (fixed-length blocks resampled with replacement) at two levels.

\paragraph{Per-point bootstrap.} For each pressure, $\phi(t)$ and $\psi_6(t)$ are sliced from \texttt{eq\_start\_index} onward (using the per-parameter cutoffs from §1: the \texttt{system.volume\_fraction} entry for $\phi$ and the \texttt{<particle>.hexatic\_6} entry for $\psi_6$, which may differ). Blocks of length $L$ frames (default $L = 50$; the agent verifies $L$ exceeds the visible autocorrelation time in the transition runs) are resampled with replacement, the equilibrated-tail mean is recomputed for each replicate, and the resampled distribution's standard deviation is reported as $\sigma(\phi)$ and $\sigma(\psi_6)$. These are the per-point error bars on $\phi(P)$ and $\psi_6(P)$.

\paragraph{Estimator-level bootstrap.} For each of $N_{\mathrm{boot}} \ge 1000$ bootstrap iterations, the agent draws a block-resampled equilibrated-tail mean independently per pressure (using the same per-parameter \texttt{eq\_start\_index} cutoffs) and recomputes $P^\ast_{\psi_6}$ and $P^\ast_\phi$ on that bootstrap replicate. The $16$/$84$ percentiles of the resulting distributions are reported as the $1$-$\sigma$ CI; the $2.5$/$97.5$ percentiles as the $95\%$ CI. Both estimators are reported with point estimate, $1$-$\sigma$ CI, and $95\%$ CI.

\paragraph{What the bootstrap does and does not capture.} The bootstrap captures within-run statistical and autocorrelation uncertainty propagated through the estimator. It does not capture:

\begin{itemize}
    \item \emph{Pressure-grid uncertainty} -- how much $P^\ast$ would shift if a different set of pressures had been sampled. This is bounded by leave-one-out: drop each point in turn (especially equilibration failures and borderline passes) and report the maximum shift in $P^\ast_{\psi_6}$ and $P^\ast_\phi$ as a sensitivity bound.
    \item \emph{Finite-$N$ bias} -- how much $P^\ast_{\psi_6}$ and $P^\ast_\phi$ would differ at larger $N$. Discussed qualitatively in caveats; quantified only if a finite-size scaling sub-sweep is performed.
    \item \emph{Residual non-equilibration} -- runs that hit the absolute cap and still fail the equilibration check are flagged, and the agent checks whether excluding them changes the result.
\end{itemize}

If the bootstrap CI looks suspiciously tight (sub-percent on $P^\ast$), the agent must say so explicitly: it reflects within-run precision, not model uncertainty.

\subsection*{5. Reporting requirements}

The final summary must contain:

\begin{enumerate}
    \item \emph{Sweep history} -- the iterations actually performed: $N$, the pressure grid added at each iteration, the \texttt{sample\_steps} actually used per point (which may vary), and the rationale for each iteration's grid choice.
    \item \emph{Equilibration-check results} per run -- the \texttt{equilibrated} flag and \texttt{eq\_start\_index} from the analysis step, with any run that was re-executed at higher \texttt{sample\_steps} or that hit the absolute cap without passing flagged explicitly.
    \item \emph{$\phi(P)$ and $\psi_6(P)$ values} with the per-point block-bootstrap $\sigma$ from §4 (table or in-line, the agent's choice).
    \item \emph{Both $P^\ast$ estimates} with point estimate, $1$-$\sigma$ CI, and $95\%$ CI from the estimator-level bootstrap (§4), with a discussion of their agreement.
    \item \emph{Sensitivity bound} -- leave-one-out shift in $P^\ast_{\psi_6}$ and $P^\ast_\phi$, and whether it exceeds the bootstrap CI.
    \item \emph{Required plots}, saved as PNG inside the simulation \texttt{working\_dir} and referenced by path:
    \begin{itemize}
        \item $\phi(P)$ with literature $\phi_L / \phi_H$ bands, $P^\ast_\phi$ marked, and per-point error bars from §4.
        \item $\psi_6(P)$ with $P^\ast_{\psi_6}$ marked and per-point error bars from §4.
        \item $g(r)$ at three pressures spanning the transition (disordered, near $P^\ast$, ordered).
        \item $\phi(t)$ traces in the transition window as a visual equilibration check.
    \end{itemize}
    \item \emph{Caveats} -- finite-size effects, equilibration uncertainty, pressure-grid sensitivity, and anything to investigate further. If the agent's $P^\ast$ differs from the literature value by more than the reported uncertainty, the agent must investigate before reporting; the most common cause is under-equilibration, addressed by re-running with more \texttt{sample\_steps}.
\end{enumerate}

\end{programbox}

The protocol's phrase ``finite-$N$ transition width'' is an instruction-level shorthand for the estimator spread the agent was asked to report. The final interpretation in Sec.~\ref{sec:autoresearch} and SI~\ref{si:autoresearch_report} treats this spread more cautiously as a combined finite-$N$ and pressure-grid sensitivity signal.

\section{Agent-Generated Research Report}
\label{si:autoresearch_report}

This SI section is a lightly edited LaTeX rendering of the final report ColPackAgent produced at the end of the autoresearch run reported in Sec.~\ref{sec:autoresearch}. The edits align notation with SI~\ref{si:autoresearch_protocol} ($\eta \to \phi$, $P^\ast_{\mathrm{OP}} \to P^\ast_{\psi_6}$, $P^\ast_\eta \to P^\ast_\phi$), correct the estimator-comparison statement discussed in Sec.~\ref{sec:autoresearch}, localize figure paths for the manuscript package, and adjust wording only where needed for consistency with those changes. We include the report because the agent's framing of caveats, sensitivity analysis, and follow-up directions, not just the numerical results, is part of the autoresearch behavior we are demonstrating.

\begin{programbox}[Agent-generated research report: 2D hard-disk freezing transition (NPT)]

\noindent\textbf{Date:} 2026-05-05.\\
\noindent\textbf{System:} 2D NPT hard disks, $N = 400$, $\sigma = 1$.\\
\noindent\textbf{Literature anchor (Bernard \& Krauth 2011):} $P^\ast \approx 9.185$, $\phi_L = 0.700$, $\phi_H = 0.716$, $\phi_{\mathrm{mid}} = 0.708$.

\subsection*{\texorpdfstring{Headline $P^\ast$ estimates}{Headline P-star estimates}}

\begin{center}
\begin{tabular}{l c c c c}
\hline
Estimator & Point & $1$-$\sigma$ CI & $95\%$ CI & LOO max shift \\
\hline
$P^\ast_{\psi_6}$ ($\psi_6$ inflection / sigmoid) & $\mathbf{8.928}$ & $[8.918, 8.936]$ & $[8.909, 8.945]$ & $0.035$ \\
$P^\ast_\phi$ ($\phi$ crosses $0.708$)            & $\mathbf{9.151}$ & $[9.150, 9.153]$ & $[9.148, 9.155]$ & $0.155$ \\
Spread $|P^\ast_{\psi_6} - P^\ast_\phi|$           & $0.224$          & ---              & ---              & --- \\
Literature $P^\ast$                                & $9.185$          & ---              & ---              & --- \\
\hline
\end{tabular}
\end{center}

Both point estimates fall below the literature value, but by different amounts. $P^\ast_\phi$ is within $0.034$ of the literature value and its LOO sensitivity range includes the literature value; $P^\ast_{\psi_6}$ is lower by $0.26$, roughly $7\times$ its LOO shift, and is consistent with the finite-$N$ rounding of the sigmoid inflection toward the liquid-to-hexatic crossover (see caveats). The spread between the two estimators is therefore treated as a finite-$N$ and pressure-grid sensitivity signal rather than as a bracket around the literature value.

\subsection*{1. Sweep history}

Each iteration is its own working directory under \texttt{data/2d\_npt\_disk*/}. Iteration 1's points at $P = 9.0, 9.5$ were superseded by iteration 3 at higher \texttt{sample\_steps}; iteration 1's solid-side points ($P = 10.5, 11.5$) were not used in the fit because iteration 2 covers the solid plateau cleanly. All other points were used.

\begin{itemize}
    \item \textbf{Iteration 1, \path{data/2d_npt_disk}, \texttt{sample\_steps}: $3\times 10^5$.}
    Pressures: $6.5$, $7.5$, $8.5$ (used); $9.0$, $9.5$, $10.5$, $11.5$ (superseded). Rationale: coarse exploratory bracket.
    \item \textbf{Iteration 2, \path{data/2d_npt_disk_v2}, \texttt{sample\_steps}: $1.5\times 10^6$.}
    Pressures: $9.7$, $9.8$, $9.9$, $10.0$, $10.1$, $10.2$, $10.3$, $10.4$. Rationale: refined solid-side sweep after iteration 1 showed the transition between $9.5$ and $10.5$.
    \item \textbf{Iteration 3, \path{data/2d_npt_disk_v3}, \texttt{sample\_steps}: $2\times 10^6$.}
    Pressures: $9.0$, $9.2$, $9.3$, $9.4$, $9.5$, $9.6$. Rationale: at iteration-2 \texttt{sample\_steps}, even $P=9.7$ was solid; the transition is below $9.7$, so the $P \approx 9$ region was rerun at refinement tier.
    \item \textbf{Iteration 4, \path{data/2d_npt_disk_v4}, \texttt{sample\_steps}: $2\times 10^6$.}
    Pressures: $8.6$, $8.8$, $8.9$, $9.1$. Rationale: anchor the rising edge; iteration 3 left a gap between fluid ($P \le 8.5$ at $3 \times 10^5$) and partially ordered ($P = 9.0$ at $2 \times 10^6$).
\end{itemize}

Total used in fit: \textbf{21 pressures}.

\subsection*{2. Equilibration check (per the program §1)}

Every used run reports \texttt{equilibrated == true} from the analysis tool's auto-cutoff for \emph{both} \texttt{system.volume\_fraction} and \texttt{disk\_0.hexatic\_6}. No run hit the absolute cap of $5 \times 10^6$; the escalation policy was applied at the \emph{sweep} level (iterations 3--4) rather than at the per-point level, since iteration-1 results showed under-equilibration was systematic in the transition window rather than confined to specific runs.

Per-parameter \texttt{eq\_start\_index} for refinement-tier runs (iters 2--4) sits at $\sim 85$--$90\%$ of the trajectory, leaving $1.5$--$3.0$\,k frames in the equilibrium tail. This is the bootstrap window.

\subsection*{\texorpdfstring{3. $\phi(P)$ and $\psi_6(P)$ per-point table}{3. phi(P) and psi6(P) per-point table}}

Errors are block-bootstrap $\sigma$ (block length $50$, $2000$ resamples) on the equilibrated tail using each run's per-parameter \texttt{eq\_start\_index}.

\begin{center}
\begin{tabular}{c c c c c c c c}
\hline
iter & $P$ & sample\_steps & $\langle\phi\rangle$ & $\sigma(\phi)$ & $\langle\psi_6\rangle$ & $\sigma(\psi_6)$ & eq? \\
\hline
1 & $6.5$  & $3.0\times 10^5$ & $0.6393$ & $0.0008$ & $0.0899$ & $0.0027$ & \checkmark \\
1 & $7.5$  & $3.0\times 10^5$ & $0.6571$ & $0.0007$ & $0.0984$ & $0.0055$ & \checkmark \\
1 & $8.5$  & $3.0\times 10^5$ & $0.6764$ & $0.0003$ & $0.1594$ & $0.0112$ & \checkmark \\
4 & $8.6$  & $2.0\times 10^6$ & $0.6838$ & $0.0004$ & $0.2009$ & $0.0092$ & \checkmark \\
4 & $8.8$  & $2.0\times 10^6$ & $0.6919$ & $0.0003$ & $0.3491$ & $0.0151$ & \checkmark \\
4 & $8.9$  & $2.0\times 10^6$ & $0.6927$ & $0.0004$ & $0.3241$ & $0.0166$ & \checkmark \\
3 & $9.0$  & $2.0\times 10^6$ & $0.7039$ & $0.0005$ & $0.5549$ & $0.0116$ & \checkmark \\
4 & $9.1$  & $2.0\times 10^6$ & $0.6999$ & $0.0003$ & $0.4768$ & $0.0118$ & \checkmark \\
3 & $9.2$  & $2.0\times 10^6$ & $0.7157$ & $0.0004$ & $0.6732$ & $0.0043$ & \checkmark \\
3 & $9.3$  & $2.0\times 10^6$ & $0.7074$ & $0.0005$ & $0.5911$ & $0.0134$ & \checkmark \\
3 & $9.4$  & $2.0\times 10^6$ & $0.7171$ & $0.0005$ & $0.6831$ & $0.0040$ & \checkmark \\
3 & $9.5$  & $2.0\times 10^6$ & $0.7179$ & $0.0003$ & $0.6812$ & $0.0032$ & \checkmark \\
3 & $9.6$  & $2.0\times 10^6$ & $0.7222$ & $0.0005$ & $0.7083$ & $0.0035$ & \checkmark \\
2 & $9.7$  & $1.5\times 10^6$ & $0.7185$ & $0.0005$ & $0.7138$ & $0.0039$ & \checkmark \\
2 & $9.8$  & $1.5\times 10^6$ & $0.7212$ & $0.0006$ & $0.7048$ & $0.0061$ & \checkmark \\
2 & $9.9$  & $1.5\times 10^6$ & $0.7276$ & $0.0007$ & $0.7487$ & $0.0037$ & \checkmark \\
2 & $10.0$ & $1.5\times 10^6$ & $0.7282$ & $0.0005$ & $0.7557$ & $0.0019$ & \checkmark \\
2 & $10.1$ & $1.5\times 10^6$ & $0.7293$ & $0.0005$ & $0.7426$ & $0.0033$ & \checkmark \\
2 & $10.2$ & $1.5\times 10^6$ & $0.7237$ & $0.0003$ & $0.7308$ & $0.0051$ & \checkmark \\
2 & $10.3$ & $1.5\times 10^6$ & $0.7289$ & $0.0004$ & $0.7352$ & $0.0025$ & \checkmark \\
2 & $10.4$ & $1.5\times 10^6$ & $0.7347$ & $0.0005$ & $0.7659$ & $0.0024$ & \checkmark \\
\hline
\end{tabular}
\end{center}

\textbf{Bracket:} $\langle\phi\rangle$ rises from $0.639$ at $P=6.5$ to $0.735$ at $P=10.4$; $\langle\psi_6\rangle$ rises from $0.090$ at $P=6.5$ to $0.766$ at $P=10.4$. Fluid anchors ($\langle\phi\rangle < \phi_L$) and solid anchors ($\langle\phi\rangle > \phi_H$) are both populated.

\subsection*{\texorpdfstring{4. $P^\ast$ with bootstrap CI (estimator-level, $2000$ replicates)}{4. P-star with bootstrap CI (estimator-level, 2000 replicates)}}

Sigmoid fit $\langle\psi_6\rangle(P) = A + B / (1 + e^{-k(P - x_0)})$ with $x_0 = P^\ast_{\psi_6}$:

\begin{itemize}
    \item $A = 0.0897$, $B = 0.6513$, $k = 4.638$, $x_0 = \mathbf{8.928}$.
    \item $1$-$\sigma$ CI: $[8.918, 8.936]$; $95\%$ CI: $[8.909, 8.945]$.
\end{itemize}

Linear interpolation for $P^\ast_\phi$ at $\langle\phi\rangle = 0.708$:

\begin{itemize}
    \item Crossing in $(P=9.0, P=9.2)$ since $\langle\phi\rangle$ runs $0.7039 \to 0.7157 \to 0.7074 \to 0.7171$. The deterministic interpolator returns the first sign change, which lands at $P^\ast_\phi = \mathbf{9.151}$.
    \item $1$-$\sigma$ CI: $[9.150, 9.153]$; $95\%$ CI: $[9.148, 9.155]$.
\end{itemize}

The two estimators do \emph{not} overlap inside their bootstrap CIs; their spread ($0.22$) far exceeds within-run uncertainty. The block bootstrap captures only autocorrelation-aware within-run noise; \textbf{the true model uncertainty is dominated by the model and grid choice (see §5/§6)}.

\subsection*{5. Sensitivity (leave-one-out)}

For each used pressure, recompute the relevant estimator with that point dropped and report the maximum shift in $P^\ast$:

\begin{itemize}
    \item \textbf{$P^\ast_{\psi_6}$}: max shift $0.035$ (dropping $P = 9.0$). LOO range $[8.893, 8.962]$, exceeding the $1$-$\sigma$ bootstrap width ($0.018$) by $\sim 2\times$ and dominating the within-run uncertainty.
    \item \textbf{$P^\ast_\phi$}: max shift $0.155$ (dropping $P = 9.0$, one endpoint of the $\phi$ interpolation interval and the sole point pulling the crossing below $9.2$). LOO range $[9.151, 9.306]$, which massively exceeds the within-run CI width ($0.003$).
\end{itemize}

$P^\ast_\phi$ is essentially set by \emph{two} points ($P = 9.0$ and $P = 9.2$), so its bootstrap CI is misleadingly tight: a different choice of grid point at $\langle\phi\rangle \approx 0.708 \pm$ a few thousandths would move the crossing by $\sim 0.15$ in $P$. The LOO bound is the honest uncertainty for $P^\ast_\phi$.

\subsection*{6. Plots}

The four plots generated with the report are bundled with this manuscript and reproduced in Fig.~\ref{fig:autoresearch_plots} below the box.

\begin{itemize}
    \item \texttt{fig\_eta\_vs\_P.png} -- $\phi(P)$ with the literature coexistence band ($\phi_L = 0.700$, $\phi_H = 0.716$), midpoint $\phi = 0.708$, $P^\ast_\phi$ marker, literature $P^\ast = 9.185$, per-point block-bootstrap error bars, and points colored by iteration.
    \item \texttt{fig\_psi6\_vs\_P.png} -- $\psi_6(P)$ with the sigmoid fit, $P^\ast_{\psi_6}$ marker, literature $P^\ast$, and per-point error bars.
    \item \texttt{fig\_rdf.png} -- $g(r)$ at three pressures ($P = 6.5$ fluid, $P = 9.2$ near transition, $P = 10.4$ solid), each averaged over the last $200$ frames of \texttt{sample\_trajectory.gsd}. Shows the expected progression: smooth liquid-like decay $\to$ split second peak $\to$ sharp crystalline peaks at $\sqrt{3}\sigma$, $2\sigma$, etc.
    \item \texttt{fig\_eta\_traces.png} -- $\phi(t)$ for the ten transition-window pressures ($P \in [8.6, 9.6]$), with each run's \texttt{eq\_start\_index} marked as a dotted vertical line. The traces show late-time stabilization in this window, with the auto-detector conservatively cutting near the $90\%$ mark to capture the stable tail, consistent with the ``passes eq-check'' verdict.
\end{itemize}

\subsection*{7. Caveats}

\begin{enumerate}
    \item \textbf{Finite-size rounding.} $N = 400$ is the high end of the program's range but small for the hexatic transition. The 2D KTHNY/Bernard--Krauth scenario predicts the inflection in $\psi_6(P)$ sits below the $\phi$-coexistence midpoint at small $N$, because the hexatic region rounds across the liquid--hexatic boundary ($P_{\mathrm{LH}} \approx 9.0$--$9.1$) and the hexatic--solid boundary ($P_{\mathrm{HS}} \approx 9.18$). At $N = 400$ those two boundaries cannot be resolved; they appear as a single, finite-width sigmoid. \textbf{Our $P^\ast_{\psi_6} = 8.928$ likely tracks the liquid-to-hexatic crossover, while $P^\ast_\phi = 9.151$ tracks the $\phi$-jump near the hexatic--solid boundary.} The literature $P^\ast = 9.185$, the \emph{coexistence} pressure for the first-order H--S transition in the thermodynamic limit, sits closer to $P^\ast_\phi$, as expected. We did not run a finite-size scaling sub-sweep to quantify this.
    \item \textbf{Pressure-grid sensitivity $>$ within-run noise.} $P^\ast_\phi$'s deterministic interpolation has a $1$-$\sigma$ bootstrap CI of $\pm 0.002$ but a LOO sensitivity of $0.15$. A modest re-binning of the $\phi$ grid would move the answer by $\sim 0.1$. The bootstrap CI should not be quoted alone; the LOO bound is the dominant uncertainty.
    \item \textbf{Equilibration tail length.} Refinement-tier runs (iters 2--4) have \texttt{eq\_start} at $\sim 88$--$90\%$ of the trajectory, leaving $1.5$--$3$\,k frames after the auto-cutoff. With block length $50$ this gives $30$--$60$ effective blocks per run. The eq-check passes everywhere, but for the transition-window runs ($P = 9.0$--$9.4$) the system spends most of its trajectory in fluctuating coexistence; the ``equilibrated'' tail is the post-nucleation regime. A run on the same point with a different RNG could converge to a different metastable state at finite $N$. We did not run replicate seeds.
    \item \textbf{Iteration-1 transition points were under-equilibrated.} At \texttt{sample\_steps}\,$= 3 \times 10^5$, the runs at $P = 9.0, 9.5$ reported \texttt{equilibrated == true} with \texttt{eq\_start} $\approx 87\%$ of the trajectory, but iteration 3 at $2 \times 10^6$ shows the same pressures \emph{had not yet crystallized} in the iter-1 window. Bernard--Krauth coexistence times scale $\propto N^2$--$N^3$; iteration 1's auto-passing under-sampled the coexistence basin. Conclusion for future studies: in 2D NPT near a first-order line, \texttt{sample\_steps} should be set to $\ge 1$--$2 \times 10^6$ from the start in the suspected transition window. \textbf{The auto-cutoff alone is not sufficient evidence of true equilibration}; cross-iteration consistency at higher steps is.
    \item \textbf{Iteration-1 solid-side points ($P = 10.5, 11.5$).} Excluded from the fit. They reported lower $\langle\phi\rangle$ than the iter-2 plateau ($0.7229$ vs.\ $0.7347$ at the comparable pressures), consistent with truncated equilibrium tails ($\sim 400$ frames). Including them would pull $P^\ast_\phi$ rightward by $\sim 0.05$, still well inside the LOO bound.
    \item \textbf{No replicate seeds.} Statistical uncertainty on each pressure is bootstrap within one trajectory, not across independent runs. For a publication-grade answer one would run $3$--$5$ replicate seeds per pressure.
\end{enumerate}

\subsection*{Files}

\begin{itemize}
    \item \texttt{pstar\_estimates.json} -- per-point table, point estimates, all bootstrap percentiles, LOO drops.
    \item \texttt{sweep\_history.json} -- iterations metadata.
    \item \texttt{trajectories.json} -- $\phi(t)$, $\psi_6(t)$ series and \texttt{eq\_start} indices, used for the trace plot.
    \item \texttt{aggregate.py} -- bootstrap pipeline (run from a \texttt{hoomd-env} shell).
    \item \texttt{make\_plots.py} -- generates the four PNGs (uses \texttt{freud} for RDFs).
    \item \texttt{fig\_eta\_vs\_P.png}, \texttt{fig\_psi6\_vs\_P.png}, \texttt{fig\_rdf.png}, \texttt{fig\_eta\_traces.png}.
\end{itemize}

Only the four PNG files are bundled in the manuscript package; the JSON tables and scripts listed above remain in the generated output directory for the run.

\end{programbox}

\begin{figure}[!h]
    \centering
    \begin{minipage}[h]{0.4\linewidth}
        \centering
        \includegraphics[width=\linewidth]{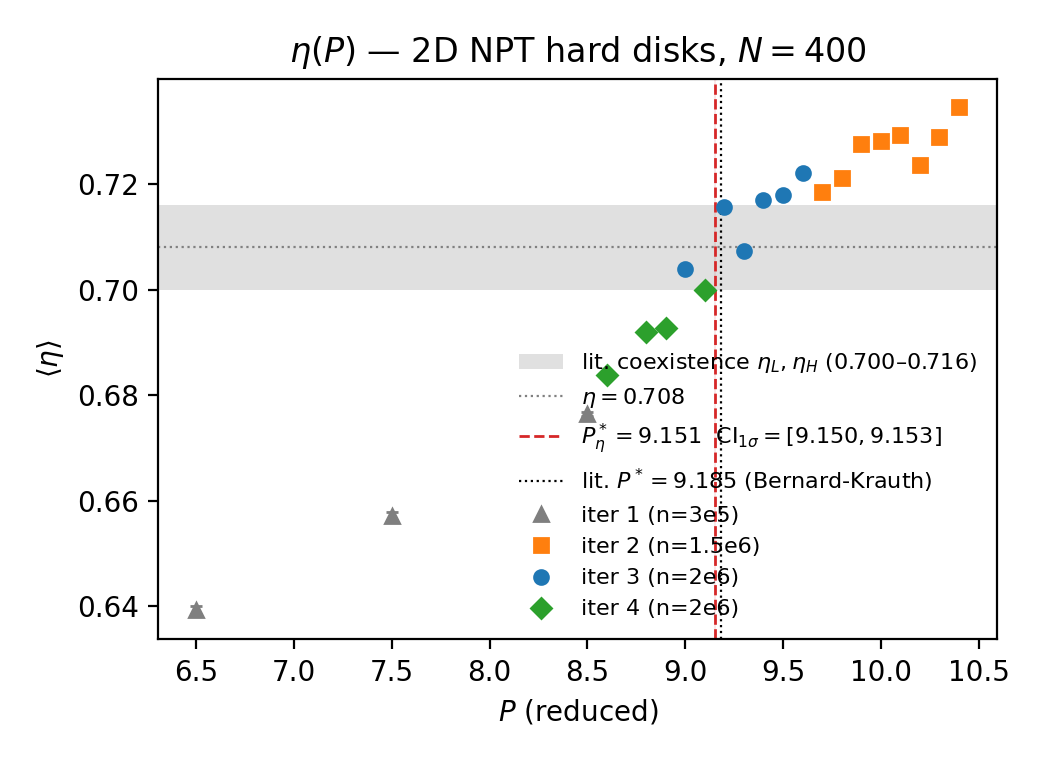}\\
        (a) $\phi(P)$
    \end{minipage}\hfill
    \begin{minipage}[h]{0.4\linewidth}
        \centering
        \includegraphics[width=\linewidth]{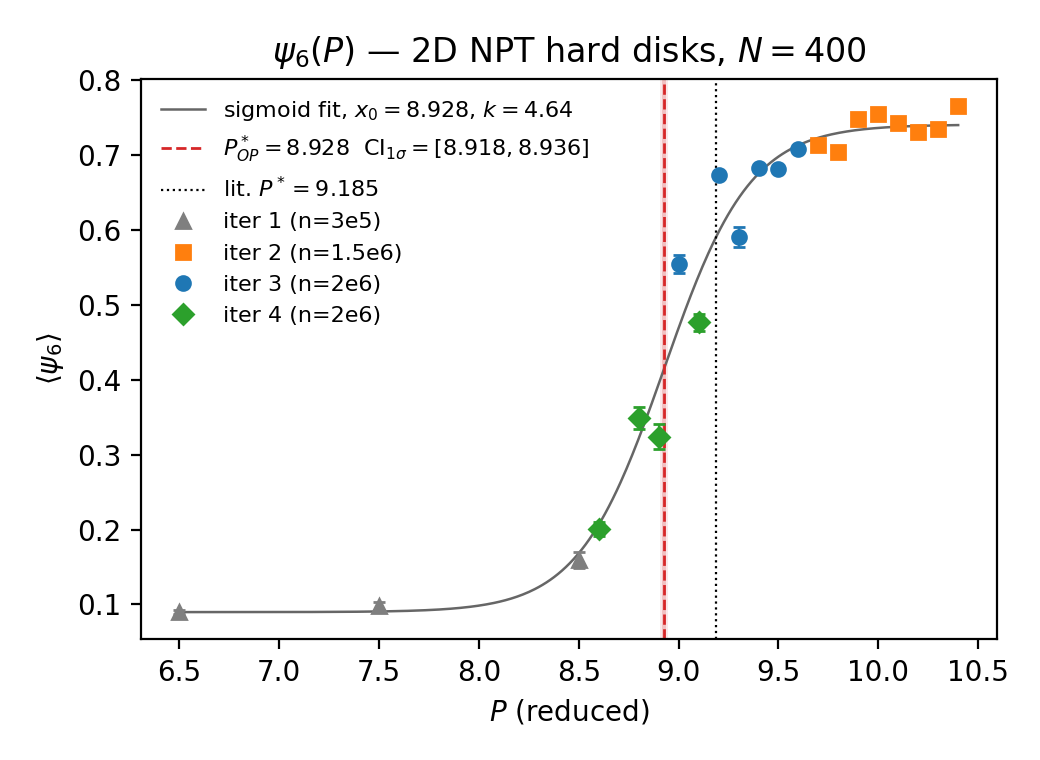}\\
        (b) $\psi_6(P)$
    \end{minipage}

    \vspace{0.5em}

    \begin{minipage}[h]{0.4\linewidth}
        \centering
        \includegraphics[width=\linewidth]{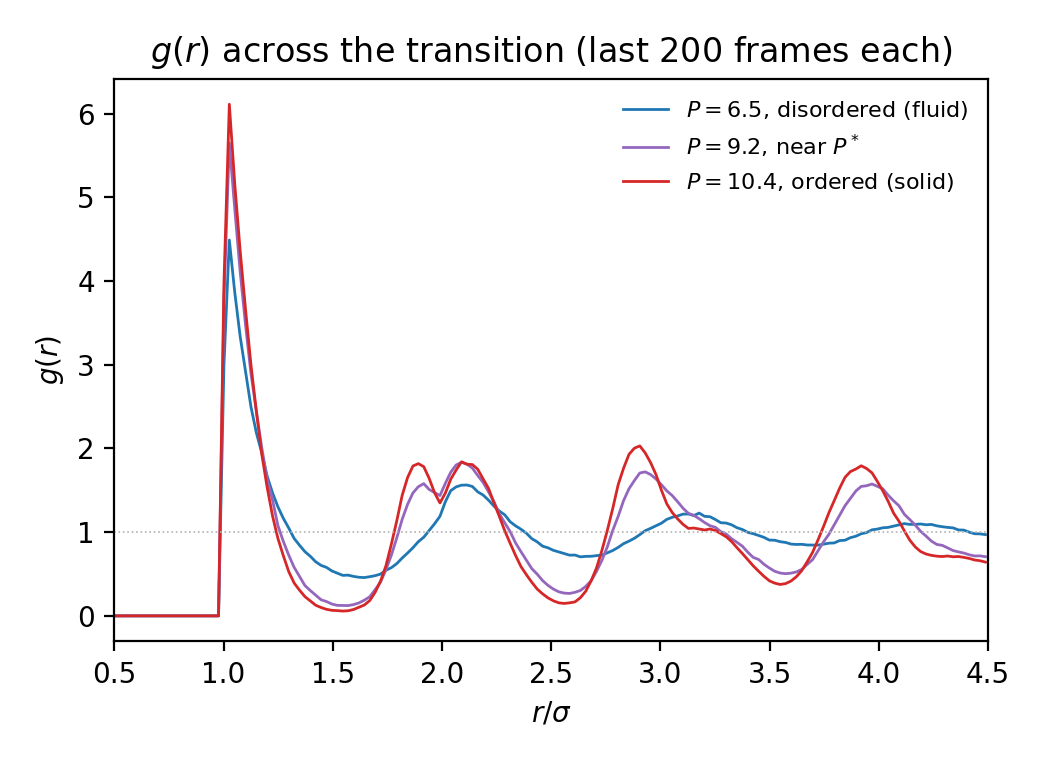}\\
        (c) $g(r)$ at three pressures
    \end{minipage}\hfill
    \begin{minipage}[h]{0.4\linewidth}
        \centering
        \includegraphics[width=\linewidth]{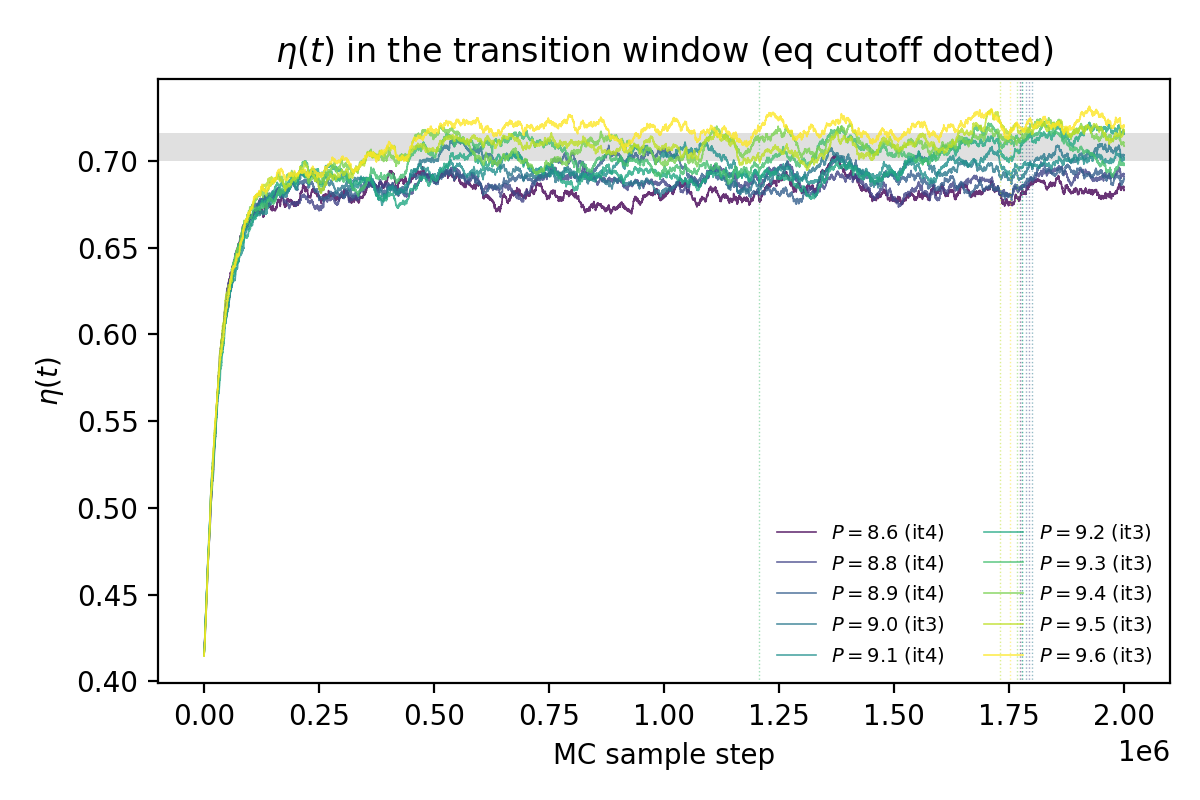}\\
        (d) $\phi(t)$ traces in the transition window
    \end{minipage}
    \caption{The four research-output plots produced by ColPackAgent at the end of the autoresearch run reported in Sec.~\ref{sec:autoresearch}. (a) $\phi(P)$ with the Bernard--Krauth coexistence band ($\phi_L = 0.700$, $\phi_H = 0.716$), midpoint $\phi = 0.708$, $P^\ast_\phi$ marker, literature $P^\ast = 9.185$, per-point block-bootstrap error bars, and points colored by sweep iteration. (b) $\psi_6(P)$ with the sigmoid fit, $P^\ast_{\psi_6}$ marker, literature $P^\ast$, and per-point error bars. (c) $g(r)$ at three pressures spanning the transition: $P = 6.5$ (fluid), $P = 9.2$ (near transition), $P = 10.4$ (solid), each averaged over the last $200$ frames. (d) $\phi(t)$ traces for the ten transition-window pressures ($P \in [8.6, 9.6]$), with each run's \texttt{eq\_start\_index} marked as a dotted vertical line.}
    \label{fig:autoresearch_plots}
\end{figure}

\section{Benchmark Task Prompts}
\label{si:benchmark_prompts}

This SI section lists the $17$ benchmark prompts evaluated in Sec.~\ref{sec:benchmark_validation}, grouped by workflow stage and ordered by difficulty. Difficulty levels $1$--$3$ are routine; levels $4$--$5$ are adversarial cases where the correct behavior is principled refusal rather than tool invocation, marked with $\dagger$. These daggered tasks are scored by transcript review against the expected refusal-or-clarification behavior rather than by expected-tool-call alone. Each prompt is followed at runtime by a stage-specific instruction to proceed without user confirmation and stop at the end of that stage; this boilerplate is identical within a stage and is omitted below for compactness. Fixture-backed prompts contain a \texttt{\{\{fixture:...\}\}} placeholder that the eval runner substitutes with a concrete working-directory path before the model sees the prompt.

\begin{programbox}[Benchmark task prompts (17 total)]

\subsection*{Setup tasks (6)}

\begin{description}
    \item[\texttt{setup\_2d\_disk} (L1)] ``Set up a 2D NVT Monte Carlo simulation of 500 hard disks.'' \emph{Expected:} call \texttt{setup\_simulation\_problem\_tool} with \texttt{dimension=2}, \texttt{ensemble=NVT}, \texttt{total\_particle\_number=500}, \texttt{particle\_shape\_list=['disk']}, then stop.

    \item[\texttt{setup\_3d\_sphere} (L1)] ``Set up a 3D NPT Monte Carlo simulation of 1000 hard spheres.'' \emph{Expected:} call \texttt{setup\_simulation\_problem\_tool} with \texttt{dimension=3}, \texttt{ensemble=NPT}, \texttt{total\_particle\_number=1000}, \texttt{particle\_shape\_list=['sphere']}, then stop.

    \item[\texttt{setup\_2d\_bidisperse\_disks} (L2)] ``Set up a 2D NVT simulation of a bidisperse mixture of disks: 500 large disks and 500 small disks (1000 total). Diameters will be set later.'' \emph{Expected:} call \texttt{setup\_simulation\_problem\_tool} with \texttt{particle\_shape\_list=['disk', 'disk']} (two entries) and \texttt{total\_particle\_number=1000}; do not invent per-component diameters at the setup stage.

    \item[\texttt{setup\_2d\_disk\_capsule} (L3)] ``Set up a 2D NPT Monte Carlo simulation with 800 particles: a mixture of disks and capsules.'' \emph{Expected:} call \texttt{setup\_simulation\_problem\_tool} with \texttt{dimension=2}, \texttt{ensemble=NPT}, \texttt{total\_particle\_number=800}, \texttt{particle\_shape\_list=['disk', 'capsule']} (order may vary).

    \item[\texttt{setup\_2d\_incompatible\_ellipse\_capsule} (L4 $\dagger$)] ``Set up a 2D NVT simulation of 600 particles: half ellipses and half capsules.'' \emph{Expected:} flag the HPMC compatibility issue (2D ellipse cannot mix with capsule/polygon shapes) and ask the user to clarify (e.g., drop the ellipse for a disk, or drop the capsule) instead of silently calling the setup tool with the incompatible mixture.

    \item[\texttt{setup\_dim\_mismatch\_disk\_cube} (L5 $\dagger$)] ``Set up a Monte Carlo simulation with 500 particles: a mixture of disks and cubes.'' \emph{Expected:} identify that disk is a 2D-only shape and cube a 3D-only shape, then ask the user for clarification rather than silently coercing the prompt into one dimension.
\end{description}

\subsection*{Planning tasks (6)}

\begin{description}
    \item[\texttt{plan\_vf\_sweep\_nvt} (L1)] ``There is an existing simulation problem already set up at the working directory \texttt{\{\{fixture:2d\_nvt\_disk\_capsule\_setup\}\}} (2D NVT, 500 particles, mixture of disks and capsules). Plan a sweep over volume\_fraction at $0.3$, $0.5$, and $0.7$.'' \emph{Expected:} call \texttt{plan\_simulation\_runs\_tool} with the fixture \texttt{working\_dir} and \texttt{tunable\_parameters=\{'volume\_fraction': [0.3, 0.5, 0.7]\}}, producing three planned runs; do not call setup.

    \item[\texttt{plan\_p\_sweep\_npt} (L1)] ``There is an existing simulation problem already set up at the working directory \texttt{\{\{fixture:2d\_npt\_disk\_capsule\_setup\}\}} (2D NPT, 500 particles, mixture of disks and capsules). Plan a pressure sweep at $P = 1.0, 5.0$, and $10.0$.'' \emph{Expected:} call \texttt{plan\_simulation\_runs\_tool} with \texttt{tunable\_parameters=\{'P': [1.0, 5.0, 10.0]\}}; three planned runs; do not call setup.

    \item[\texttt{plan\_diameter\_sweep\_bidisperse} (L2)] ``There is an existing simulation problem set up at \texttt{\{\{fixture:2d\_nvt\_disk\_disk\_setup\}\}} (2D NVT, 1000 disks as a binary mixture of two disk types). Use volume\_fraction = 0.5 as the baseline. Plan a sweep over the first disk type's diameter at $0.5$, $1.0$, $1.5$, with the second disk type fixed at diameter $1.0$.'' \emph{Expected:} call \texttt{plan\_simulation\_runs\_tool} with \texttt{baseline\_parameters} including \texttt{'volume\_fraction':\,0.5} and \texttt{'particle\_specs.1.diameter':\,1.0}, and \texttt{tunable\_parameters=\{'particle\_specs.0.diameter': [0.5, 1.0, 1.5]\}}; three planned runs; do not call setup.

    \item[\texttt{plan\_relative\_vf\_sweep} (L3)] ``There is an existing simulation problem at \texttt{\{\{fixture:2d\_nvt\_disk\_capsule\_setup\}\}} (2D NVT, 500 particles, mixture of disks (component 0) and capsules (component 1)). Use volume\_fraction = 0.4 as the baseline. Plan a sweep where the capsule volume fraction relative to disks is $0.5$, $1.0$, and $2.0$. The disk component stays as the reference (relative\_volume\_fraction = 1).'' \emph{Expected:} call \texttt{plan\_simulation\_runs\_tool} with \texttt{tunable\_parameters=\{'particle\_specs.1.relative\_volume\_fraction': [0.5, 1.0, 2.0]\}}, the disk component's relative volume fraction left at $1$; three planned runs; do not call setup.

    \item[\texttt{plan\_two\_axis\_sweep} (L3)] ``There is an existing simulation problem at \texttt{\{\{fixture:2d\_nvt\_disk\_capsule\_setup\}\}} (2D NVT, 500 particles, disks and capsules). Use volume\_fraction = 0.5 baseline. Plan two separate sweeps: volume\_fraction at $0.3$, $0.5$, $0.7$, and capsule length at $2.0$, $3.0$, $4.0$. Each varied independently from the baseline.'' \emph{Expected:} call \texttt{plan\_simulation\_runs\_tool} once with \texttt{tunable\_parameters} containing \emph{both} \texttt{'volume\_fraction': [0.3, 0.5, 0.7]} and \texttt{'particle\_specs.1.length': [2.0, 3.0, 4.0]}, producing six planned runs (3 + 3, not 9 from a Cartesian product).

    \item[\texttt{plan\_invalid\_p\_on\_nvt} (L4 $\dagger$)] ``There is an existing simulation problem at \texttt{\{\{fixture:2d\_nvt\_disk\_capsule\_setup\}\}} (2D NVT, 500 particles, disks and capsules). Plan a pressure sweep at $P = 1.0, 5.0$, and $10.0$.'' \emph{Expected:} identify that pressure is not a valid sweep parameter under NVT; either ask the user to switch the ensemble to NPT or refuse to call the planner with an invalid path, instead of silently submitting $P$ as a tunable on an NVT problem.
\end{description}

\subsection*{Analysis tasks (5)}

\begin{description}
    \item[\texttt{analyze\_2d\_disks\_default} (L1)] ``An existing simulation has already been run end-to-end at the working directory \texttt{\{\{fixture:2d\_nvt\_disk\}\}} (2D NVT, 100 hard disks, volume\_fraction sweep at $0.3, 0.5, 0.7, 0.8$). Run the analysis with default parameters.'' \emph{Expected:} call \texttt{analyze\_simulation\_runs\_tool} with the fixture \texttt{working\_dir} and no \texttt{extra\_order\_params}; per-shape defaults (Hexatic $\psi_6$ for disks) apply; surface the $\psi_6$ values across all four runs; do not call setup or plan.

    \item[\texttt{analyze\_2d\_disks\_interpret} (L2)] ``An existing simulation has already been run end-to-end at the working directory \texttt{\{\{fixture:2d\_nvt\_disk\}\}} (2D NVT, 100 hard disks, volume\_fraction sweep at $0.3, 0.5, 0.7, 0.8$). Run the analysis with default parameters and tell me at which volume fraction the disks start to show clear hexagonal order. Use the Hexatic $\psi_6$ trend across the sweep to back up your answer.'' \emph{Expected:} run default analysis; observe a sharp $\psi_6$ increase as $\phi \to 0.7$ and identify that volume fraction in the reply, grounded in the numerical results rather than generic physics knowledge.

    \item[\texttt{analyze\_2d\_disks\_extra\_hexatic} (L3)] ``An existing simulation has been run at \texttt{\{\{fixture:2d\_nvt\_disk\}\}} (2D NVT, 100 hard disks, volume\_fraction sweep). In addition to the default Hexatic $\psi_6$, also compute Hexatic $\psi_4$ (call it \texttt{hexatic\_4}) and Hexatic $\psi_3$ (\texttt{hexatic\_3}), so I can check for square and triangular ordering signatures alongside the hexagonal one.'' \emph{Expected:} call \texttt{analyze\_simulation\_runs\_tool} with \texttt{extra\_order\_params=[\{name: 'hexatic\_4', type: 'Hexatic', params: \{k: 4\}\}, \{name: 'hexatic\_3', type: 'Hexatic', params: \{k: 3\}\}]} (order may vary); do not duplicate the default $\psi_6$ in extras; reply surfaces $\psi_3$, $\psi_4$, $\psi_6$ across the sweep.

    \item[\texttt{analyze\_2d\_mixture\_per\_shape} (L4)] ``An existing simulation has been run at \texttt{\{\{fixture:2d\_nvt\_disk\_capsule\}\}} (2D NVT, 100 particles, 50 disks + 50 capsules, volume\_fraction sweep at $0.4$ and $0.6$). Analyze the runs. The defaults already cover Hexatic $\psi_6$ for disks and Nematic for capsules --- keep those. Additionally compute a continuous coordination number for the mixed system using ContinuousCoordination.'' \emph{Expected:} call \texttt{analyze\_simulation\_runs\_tool} with a system-wide \texttt{ContinuousCoordination} entry in \texttt{extra\_order\_params}; per-shape defaults (Hexatic $\psi_6$ for disks, Nematic for capsules) still appear in the results; reply summarizes both per-shape order and the continuous coordination across the two $\phi$ runs.

    \item[\texttt{analyze\_2d\_disks\_cubatic\_adversarial} (L5 $\dagger$)] ``An existing simulation has been run at \texttt{\{\{fixture:2d\_nvt\_disk\}\}} (2D NVT, 100 hard disks, volume\_fraction sweep). Analyze the runs and include a Cubatic order parameter in the analysis.'' \emph{Expected:} flag that Cubatic measures cubic symmetry from per-particle orientation vectors and is meaningless for 2D isotropic disks; ask the user to clarify (e.g., did they mean Hexatic $\psi_4$ for square ordering, or did they mean to set up a 3D cube system?) rather than blindly forwarding ``Cubatic'' to the analyze tool.
\end{description}

\end{programbox}

\end{document}